\documentclass[11pt]{article}

\usepackage[utf8]{inputenc}
\usepackage[T1]{fontenc}
\usepackage{geometry}
\usepackage{graphicx}
\usepackage{amsmath,amssymb}
\usepackage[numbers,sort&compress]{natbib}
\usepackage{booktabs}
\usepackage{tabularx}
\usepackage{array}
\usepackage{xcolor}
\usepackage{enumitem}
\usepackage{float}
\usepackage{titlesec}
\usepackage{microtype}
\usepackage[hyphens]{url}
\usepackage{tikz}
\usetikzlibrary{arrows.meta,positioning}
\usepackage[font=small,labelfont=bf]{caption}
\usepackage{placeins}
\usepackage[breaklinks,colorlinks=true,linkcolor=blue,citecolor=blue,urlcolor=blue]{hyperref}

\titlespacing*{\section}{0pt}{12pt plus 4pt minus 2pt}{6pt plus 2pt minus 2pt}
\titlespacing*{\subsection}{0pt}{10pt plus 3pt minus 2pt}{4pt plus 2pt minus 2pt}
\titlespacing*{\subsubsection}{0pt}{8pt plus 2pt minus 2pt}{3pt plus 1pt minus 1pt}

\definecolor{figNavy}{HTML}{264653}
\colorlet{figMono}{figNavy}

\hypersetup{
  pdfauthor={Majid Memari, George Rudolph},
  pdftitle={The Capability Ladder: A Curriculum-Modernization Framework for Workforce Readiness in the AI Era}
}

\title{\textbf{The Capability Ladder: A Curriculum-Modernization Framework for Workforce Readiness in the AI Era}}

\author{
  Majid Memari\textsuperscript{1,*}\thanks{ORCID: 0000-0001-5654-4996} \and
  George Rudolph\textsuperscript{1}\\[0.5em]
  \small \textsuperscript{1}Department of Computer Science, Utah Valley University, Orem, UT 84058, USA\\[0.5em]
  \small *Correspondence: mmemari@uvu.edu
}

\date{}

\begin{document}

\maketitle

\begin{abstract}
Artificial intelligence is changing the task composition of computing work faster than curricula and training typically adapt. This is a curriculum-framework paper, grounded in a structured narrative review of labor-market and software-engineering evidence and illustrated through an exploratory pilot course: the review supports the framework, and the pilot illustrates it rather than serving as primary evidence. The central claim is that near-term change is task reallocation rather than full replacement: routine implementation is increasingly automated while verification, systems thinking, security, and the ability to \emph{supervise and orchestrate AI}---keeping a human in the loop---gain value. We organize the response as a capability-assurance framework anchored by a \emph{Capability Ladder}: a five-level progression (trigger, automation, workflow, AI agent, agent team) that classifies the operational autonomy of AI-augmented work and the human supervision it requires. We map the ladder to course-level updates, workload-aware assessment, and stackable workforce credentials, and illustrate it through a two-semester pilot of a team-based, no-code course enrolling computing and business students. We argue for \emph{targeted modernization} around durable capabilities rather than wholesale curriculum replacement, and we are explicit about evidence limits: labor signals are confounded by non-AI forces, industry reports are directional, and the pilot is exploratory.
\end{abstract}


\section{Introduction}
AI-augmented tools are changing what entry-level and mid-level computing professionals do each day: AI can accelerate drafting and coding, but it also increases the need for validation, testing, and contextual judgment \cite{noy2023,spinellis2024,cui2024}. Labor-market studies show growth in AI-related demand and shifting skill composition within roles, rather than immediate elimination of whole occupations \cite{acemoglu2022,manca2023,borgonovi2023}, while computing-education syntheses describe a parallel inflection point in which traditional assignment and assessment designs no longer suffice to measure individual competence \cite{denny2024,iticsewg2023,sengul2024,shein2024}. Computing education has long faced a lag between classroom tools and workplace practice; generative AI intensifies this familiar problem by accelerating how fast professional workflows change, for degree-seeking students and practicing professionals alike. Institutions have responded unevenly---some rebuilding courses around whichever tools are newest, others holding to pre-AI curricula---and individual instructors are already revising assessment and course design around AI \cite{gross2026university}. Yet neither extreme serves students: over-reaction chases volatile tool fads, while under-reaction leaves graduates behind current practice. A further risk cuts the other way---in the rush to adapt, programs can lose the durable computing foundations that make any tool usable. What is clear is that AI is changing computing work and, with it, computing education; what is contested is \emph{how} programs should respond, and how much.

Teaching-focused institutions are unusually well positioned to serve both fronts, because they routinely operate degree programs, continuing education, and employer partnerships side by side. This paper therefore asks a practical question for teaching-focused computing programs (including the Consortium for Computing Sciences in Colleges audience): how can a single, evidence-based competency model prepare \emph{both students and professionals} for AI-augmented computing roles---proportionately, neither over- nor under-reacting? \textbf{This is a curriculum-framework paper, grounded in a structured narrative review and illustrated through an exploratory pilot course}: the review supports the framework, and the pilot illustrates it rather than serving as the main evidence.

Our contributions are fourfold:
\begin{enumerate}[noitemsep,topsep=2pt]
  \item We synthesize recent labor-market and computing-education evidence to identify which capabilities are becoming more important in AI-augmented computing work.
  \item We propose the \emph{Capability Ladder} as a practical model for describing increasing levels of AI autonomy and human supervision.
  \item We map the ladder to course-level updates, assessment strategies, and workforce-upskilling pathways.
  \item We illustrate the model through an exploratory two-semester pilot course involving computing and business students.
\end{enumerate}

The paper proceeds as follows: Section~\ref{sec:method}, review method and evidence boundaries; Section~\ref{sec:framing}, conceptual framing; Section~\ref{sec:evidence}, evidence on AI, work, and computing skills; Section~\ref{sec:ladder}, the Capability Ladder framework; Section~\ref{sec:courses}, course-level implementation; Section~\ref{sec:assessment}, workload-aware assessment; Section~\ref{sec:workforce}, the workforce and stackable-credential pathway; Section~\ref{sec:pilot}, an exploratory pilot course; Section~\ref{sec:roadmap}, a staged adoption roadmap; Section~\ref{sec:limits}, limitations and risks; and Section~\ref{sec:conclusion}, conclusion.

\section{Review method and evidence boundaries}\label{sec:method}
This paper uses a structured narrative review designed for curriculum decision support rather than meta-analysis. The protocol (1)~framed questions around entry-level computing-task change, AI-skill demand diffusion, and computing-assessment validity under AI availability; (2)~prioritized post-2023 studies, adding selected pre-2023 labor-framing work to define exposure, augmentation, and displacement; (3)~screened for empirical labor-market signals, controlled/field evidence on AI productivity, and peer-reviewed computing-education syntheses; and (4)~weighted each source by an evidence-confidence tier rather than by citation count.

\paragraph{Search process.} We searched the ACM Digital Library, IEEE Xplore, and Google Scholar for peer-reviewed work, and consulted institutional and industry sources---the OECD, the U.S.\ Bureau of Labor Statistics (BLS), the World Economic Forum (WEF), and selected analyst or vendor reports---for labor-market signals. Representative search terms included \emph{generative AI and computing education}, \emph{AI skills demand}, \emph{AI and software-engineering productivity}, \emph{AI agents and human supervision}, \emph{AI assessment redesign}, and \emph{workforce upskilling AI}, combined with computing-specific qualifiers. This was a purposive, decision-oriented search rather than an exhaustive systematic review.

\paragraph{Inclusion and exclusion criteria.} We preferred post-2023 sources; prioritized peer-reviewed studies where available; included official labor-market reports; used industry reports only as directional evidence; and admitted pre-2023 sources only for baseline concepts (exposure, automation, augmentation, displacement). We excluded opinion pieces without evidence, vendor marketing material, sources focused solely on K--12 education (our scope is college and workforce readiness), and studies unconnected to computing, AI skills, or curriculum design.

\paragraph{Evidence-confidence tiers.} Rather than treating all citations equally, we sort sources into three tiers (Table~\ref{tab:tiers}) and let tier govern how strongly a source can support a claim.

\begin{table}[ht]
\centering
\caption{Evidence-confidence tiers used to weight sources.}
\label{tab:tiers}
\small
\begin{tabular}{@{}p{0.16\linewidth}p{0.42\linewidth}p{0.34\linewidth}@{}}
\toprule
\textbf{Tier} & \textbf{Source types} & \textbf{Used for} \\
\midrule
High & Randomized and field experiments, official labor statistics, peer-reviewed studies with strong methods & Core claims about task-level productivity, assessment validity, and labor trends \\
\addlinespace[2pt]
Moderate & Large-scale job-posting studies, cross-country analyses, systematic or scoping reviews & Curriculum priorities and skill-demand signals \\
\addlinespace[2pt]
Exploratory & Industry reports, course case studies, perception studies, instructor reflections, early pilot data & Design hypotheses, examples, and implementation ideas \\
\bottomrule
\end{tabular}
\end{table}

\paragraph{Temporal scope.} We center the synthesis on 2023--2026 for a specific reason: the public release of ChatGPT in late 2022 marks the inflection after which generative AI became broadly available in both workplaces and classrooms, so 2023 is the first full year of widespread exposure. \emph{2026 sources are used only where year-to-date data were available at the time of writing, and are interpreted cautiously because the year was incomplete}; we do not treat partial-year figures as full-year trends. Selected pre-2023 sources supply a pre-GenAI baseline and the exposure/augmentation/displacement vocabulary. Industry trackers such as Layoffs.fyi are read as \emph{directional industry indicators, not causal evidence}, throughout. Because this window also overlaps pandemic-era labor volatility, we separate that disruption from GenAI effects when interpreting post-2023 signals (see the attribution analysis in Section~\ref{sec:evidence}).

The pilot evidence reported later is exploratory---a single cross-disciplinary course used to illustrate feasibility and generate design hypotheses, not to claim causal placement effects. More broadly, this is a decision-oriented synthesis for departments facing near-term revision constraints, not a comprehensive census or a source of pooled causal estimates.

\section{Conceptual framing}\label{sec:framing}
We distinguish four often-conflated terms: \textbf{exposure} (an occupation relies on abilities AI is advancing), \textbf{automation} (AI substitutes for specific tasks), \textbf{augmentation} (AI complements people and raises productivity on tasks), and \textbf{displacement} (net labor-demand reduction after substitution and productivity effects combine).

Exposure is not equivalent to displacement. Occupational-exposure studies and cross-country analyses suggest mixed effects, with stronger complementarity where digital skill depth is higher \cite{felten2021,georgieffhyee2022}. Global analyses similarly conclude that the dominant near-term channel for generative AI is augmentation of many occupations rather than full automation of entire roles \cite{gmyrek2023}. This matters for curriculum design: programs should train students for high-complementarity task bundles. This \emph{task-reallocation chain}---from AI capability progress, to task-level effects (routine drafting compresses while verification and orchestration expand), to work-role recomposition, to curriculum response---organizes the rest of the paper.

\section{Evidence on AI, work, and computing skills}\label{sec:evidence}
\paragraph{Software work is being reorganized, not eliminated.}
Three converging signals shape curriculum priorities. First, controlled and field studies show productivity gains from generative AI in selected tasks alongside reliability and overreliance concerns \cite{noy2023,cui2024}. Because AI lowers the cost of first-draft implementation, student value shifts toward creative problem framing, solution design, verification, and explanation---the focus of the assessment-redesign patterns developed below. Second, online labor-market studies report short-term demand compression in automation-prone task categories \cite{demirci2025,hui2024,teutloff2025}, implying that entry-level readiness should emphasize transferable capabilities beyond routine code production. Third, OECD vacancy analyses show AI-skill demand diffusing faster than average skills while absolute AI posting shares remain modest \cite{manca2023,borgonovi2023}, so programs should give all students baseline AI fluency while preserving advanced specialization pathways.

\paragraph{Recent labor-market indicators (2023--2026).}
From 2023 to 2026, the IT labor market shows \emph{restructuring rather than uniform collapse}: tech-sector layoffs stayed high in 2024--2025 (about 153k and 124k) before easing in 2026 \cite{layoffsfyi2026}---a directional industry indicator, not causal evidence---but the deeper pattern is recomposition. Industry analysis reports AI-skill postings growing while total postings fell, a rising AI-skill wage premium (25\% to 56\%), faster skill change in AI-exposed roles, and a widening productivity gap between the most and least AI-exposed industries (27\% vs.\ 9\%) \cite{pwc2025barometer}; as single-source figures, we read these as directional, not causal. The combined signal is consistent with rising demand for professionals who can verify AI output, orchestrate multi-step workflows, and adapt quickly to tool changes.

\paragraph{Causal attribution: AI-driven or confounded?}
A central inference risk is \emph{confounding}: several non-AI forces co-occur with the contraction in entry-level technology hiring, so attributing the whole shift to AI mistakes correlation for causation. On the demand side, post-2022 macroeconomic tightening cooled speculative hiring, a correction of pandemic-era (2020--2022) over-hiring drove much of the layoff wave, and the changed U.S.\ tax treatment of software R\&D (Section~174 amortization, effective tax-year 2022) raised the after-tax cost of engineering headcount. On the supply side, slowing working-age-population and labor-force growth shifts net hiring independently of automation, while remote work and offshoring relocate where entry-level tasks are performed \cite{blsmlr2024}. Because these factors move \emph{together} with AI adoption, observational labor data (job postings, wage series) identify \emph{association}, not the isolated causal effect of AI; consistently, the WEF employer survey treats AI as one driver among macroeconomic, demographic, and sectoral factors behind a projected 170M jobs created and 92M displaced by 2030 \cite{wef2025futurejobs}. We therefore make causal claims only where the study design supports them---chiefly at the \emph{task} level, from randomized and field experiments on AI-assisted work \cite{noy2023,cui2024}---and treat the occupation-level decline as AI-associated but not cleanly AI-attributable.

This attribution ambiguity strengthens rather than weakens the curricular stance: a program that trains durable, transferable capabilities (verification, systems thinking, security, supervision) is robust across \emph{all} of these explanations, whereas a narrowly AI-reactive redesign is not. It also argues for a proportional response---adjusting effort to signal strength rather than to headlines---which we make concrete in the roadmap of Section~\ref{sec:roadmap}.

\subsection{Emerging roles and the skills they demand}
As tasks reallocate, computing roles recompose rather than vanish. U.S.\ projections (2023--2033) show strong growth for data scientists ($+36\%$) and information-security analysts ($+33\%$), modest growth for software developers ($+18\%$), and decline for routine computer-programming roles ($-10\%$) \cite{blsmlr2024}; alongside these, AI-specific roles---machine-learning and MLOps engineers, AI product and project managers, and AI-augmented developers---are emerging, and AI-skill expectations now extend well beyond specialist positions. Studies of these roles converge on a profile pairing technical AI fluency (data handling, model use, evaluation) with \emph{durable} capabilities---verification, creative problem-solving, communication, and ethics---yet curricula still under-weight the latter relative to demand \cite{mol2024,bobitan2024}. A distinctively new competency is \emph{human--AI teaming}: structuring work so people and AI agents share tasks under clear roles and human oversight \cite{berretta2023haiteaming}. Reskilling is also uneven---not all professionals want to or can retrain effectively---so pathways must accommodate different starting points. Preparing students therefore means teaching not only how to build and use AI but how to \emph{supervise} it, and the elevated skills map cleanly onto the ladder levels defined next: AI-augmented developers work at L1--L3 (prompt and tool use, verification, debugging AI output); ML/MLOps engineers at L2--L3 (data pipelines, evaluation, deployment, monitoring); AI product and project managers at L3--L4 (scoping, orchestration, human--AI teaming, communication); and security/QA roles at L3 with an emphasis on verification (threat modeling, misuse analysis, output assurance).

\section{The Capability Ladder framework}\label{sec:ladder}
We propose a \emph{capability-assurance} model---a set of durable capabilities paired with a cross-cutting assurance layer (verification, security, documentation, ethics, and human approval) that applies whenever AI is used. It rests on five priorities:
\begin{enumerate}[noitemsep]
  \item \textbf{AI-assisted workflow competence}: deliberate and documented use of AI assistants.
  \item \textbf{Verification-first engineering}: testing, debugging, code review, and security checks.
  \item \textbf{Systems thinking and solution design}: architecture, integration, and tradeoff reasoning.
  \item \textbf{Communication and defense}: written and oral justification of technical choices.
  \item \textbf{Ethics and governance}: attribution, privacy, policy compliance, and responsible use.
\end{enumerate}

Each priority maps to an observable, employer-relevant outcome (e.g., reproducible use logs, strong test suites, clear technical defense). To operationalize them we recommend a three-strand teaching model rather than a single ``AI course'':
\begin{enumerate}[noitemsep]
  \item \textbf{Build AI foundations}: CS pathways include implementation of selected ML/deep-learning components; business and workforce pathways emphasize conceptual foundations, model limits, and error sources.
  \item \textbf{Use AI as an engineering tool}: learners practice AI-assisted coding, code generation, and retrieval/evaluation workflows with explicit verification and documentation requirements.
  \item \textbf{Supervise and orchestrate AI agents}: learners assign bounded tasks (e.g., draft tests, code-review suggestions, refactoring proposals), evaluate outputs, and retain human accountability for decisions.
\end{enumerate}
CS majors may traverse the full ``build'' strand, while business and workforce learners typically emphasize the ladder's upper levels.

\begin{figure}[t]
\centering
\begin{tikzpicture}[
  font=\small,
  rung/.style={draw, rounded corners=2pt, minimum height=6.5mm, align=left, text width=0.86\linewidth, inner sep=3pt},
  node distance=2.4mm]
\node[rung, fill=figMono!10] (l0) {\textbf{L0 Trigger:} an event starts a process (form, schedule, webhook).};
\node[rung, fill=figMono!14, above=of l0] (l1) {\textbf{L1 Automation:} one action fires from a trigger (if-this-then-that).};
\node[rung, fill=figMono!18, above=of l1] (l2) {\textbf{L2 Workflow:} fixed multi-step process across tools (validate, route, notify).};
\node[rung, fill=figMono!24, above=of l2] (l3) {\textbf{L3 AI agent:} a goal-directed system that plans and acts, \emph{with human verification}.};
\node[rung, fill=figMono!30, above=of l3] (l4) {\textbf{L4 Agent team:} multiple agents delegate and report under human supervision.};
\node[rung, fill=figMono!6, draw=figMono!55, dashed, below=of l0] (assur) {\textbf{Assurance (every level):} verification, security, documentation, ethics, human approval.};
\draw[-{Stealth[length=1.8mm]}, thick] ([xshift=-2mm]l0.south west) -- ([xshift=-2mm]l4.north west) node[midway, left, rotate=90, anchor=south, font=\scriptsize] {increasing autonomy + supervision};
\end{tikzpicture}
\caption{The Capability Ladder: five levels of operational autonomy (L0--L4), each requiring more human supervision, with a cross-cutting assurance layer applied at every level. Higher levels are reachable with low/no-code tools, lowering the barrier to participating in AI-augmented work.}
\label{fig:capabilityladder}
\end{figure}
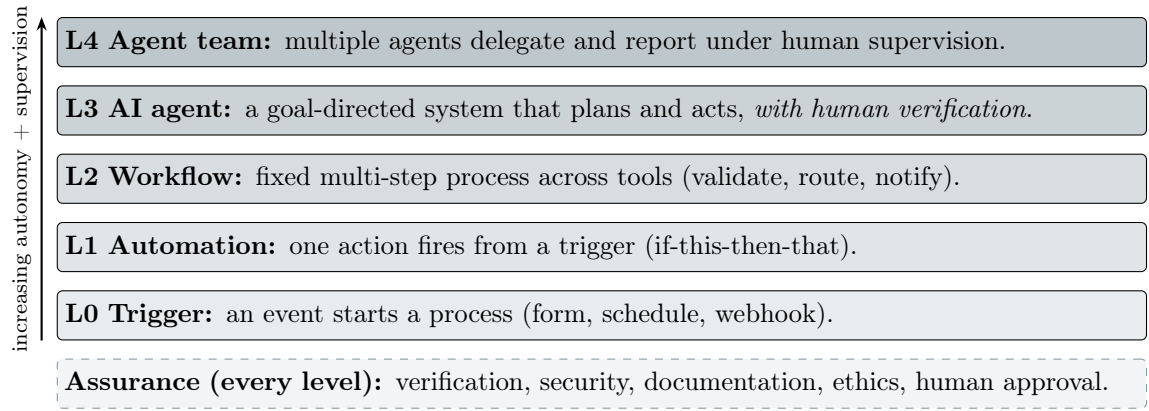

\paragraph{The Capability Ladder.} To make the ``use'' and ``supervise'' strands concrete and assessable, we define a five-level \emph{Capability Ladder} (Figure~\ref{fig:capabilityladder}). Its vertical axis is not intellectual maturity but \emph{operational autonomy paired with rising human-supervision responsibility}: each level hands more execution to AI and demands more oversight. Strands and levels are orthogonal: the strands name \emph{what} learners do with AI (build, use, supervise), while the levels mark \emph{how much} autonomy a system has and how much oversight it requires. The ladder is therefore not a universal sequence every learner must climb identically; it is a scale for describing the autonomy of the AI-enabled systems a learner can responsibly use or supervise. A cross-cutting assurance layer applies at every level, and the agent levels (L3--L4) align with emerging LLM-agent frameworks \cite{wang2024agents}.

\paragraph{What distinguishes the ladder from existing frameworks.} Related frameworks operate at other units of analysis. Institution-level frameworks---for example, the framework for AI in business education developed by Inspire Higher Ed with AACSB, the Graduate Business Curriculum Roundtable, and GMAC---identify themes leaders should coordinate, such as AI ecosystems, faculty development, and ethics \cite{means2026gmac}. Cognitive taxonomies such as Bloom's classify the complexity of \emph{thinking} \cite{andersonkrathwohl2001}, and agent taxonomies classify \emph{system architectures} \cite{wang2024agents}. The ladder's unit of analysis is different: the \emph{artifact a learner is trusted to build and supervise}. Each level names an observable class of artifact (trigger, automation, workflow, agent, agent team) together with the supervision evidence it requires, so the ladder attaches directly to individual assignments and rubrics. That assignment-level granularity---what a student may delegate, must verify, and must document---is the specific gap the ladder fills. Because higher levels are reachable on low/no-code platforms, the ladder also lowers the barrier to \emph{participating in AI-augmented work}, widening access for non-CS learners and supporting credit for prior learning.

\paragraph{Levels, examples, and assessment evidence.} Table~\ref{tab:ladderevidence} makes each level concrete: a representative student task, the evidence a student submits, and what that evidence is meant to assess. Defining the evidence alongside the level keeps the ladder usable as an assessment scaffold, not only a description.

\begin{table}[ht]
\centering
\caption{The Capability Ladder with concrete examples and the student evidence it warrants.}
\label{tab:ladderevidence}
\small
\begin{tabular}{@{}p{0.13\linewidth}p{0.27\linewidth}p{0.30\linewidth}p{0.22\linewidth}@{}}
\toprule
\textbf{Level} & \textbf{Example task} & \textbf{Student evidence} & \textbf{Assessment focus} \\
\midrule
L0 Trigger & Create a form submission or calendar event that starts a process & Trigger description; input/output screenshot; explanation of when it fires & Can the student identify the event that starts the process? \\
\addlinespace[2pt]
L1 Automation & Send a confirmation email automatically when a form is submitted & Automation rule; a test case; before/after result & Can the student explain the condition, action, and failure case? \\
\addlinespace[2pt]
L2 Workflow & Collect data, validate it, route it to the right person, and log the result & Workflow diagram; validation logic; test results; error handling & Can the student manage a reliable multi-step process? \\
\addlinespace[2pt]
L3 AI agent & Configure an agent to summarize requests, suggest next steps, and flag uncertain cases for review & Goal description; prompt/configuration; sample outputs; verification checklist; human-approval point & Can the student supervise AI output and catch errors? \\
\addlinespace[2pt]
L4 Agent team & Design a system where one agent gathers information, another drafts, another checks quality, and a human approves & Agent role map; delegation structure; audit log; risk analysis; final human-decision report & Can the student coordinate multiple AI components while maintaining accountability? \\
\bottomrule
\end{tabular}
\end{table}

Across a four-year pathway the strands deepen with the ladder: introductory courses pair modeling intuition with disclosed, human-approved AI assistance (L0--L1); core courses add ML implementation, verification checklists, and bounded agent tasks (L2); advanced courses add error diagnosis and supervised workflow integration (L3); and the capstone adds production tradeoffs and supervised multi-agent orchestration with audit reporting (L4). The same ladder also scopes shorter, stackable units for workforce learners (Section~\ref{sec:workforce}), so departments phase the model into existing courses rather than adding new ones.

\section{Course-level implementation}\label{sec:courses}
The framework maps onto existing courses without adding new ones. Table~\ref{tab:courseupdates} gives one concrete, AI-allowed-but-verified assignment per course type; in each case students may use AI, but credit depends on evidence of judgment, verification, and ownership rather than on the generated artifact alone.

\begin{table}[ht]
\centering
\caption{Concrete course-level assignment examples that allow AI but require verification.}
\label{tab:courseupdates}
\small
\begin{tabular}{@{}p{0.22\linewidth}p{0.72\linewidth}@{}}
\toprule
\textbf{Course} & \textbf{Example assignment} \\
\midrule
Intro programming & Students may use AI to generate first-draft code, but must submit the prompt log, explain two AI-generated errors, write their own tests, and complete a short live debugging task. \\
\addlinespace[2pt]
Software engineering & Teams use AI for code-review suggestions, but must decide which suggestions to accept, justify rejections, and run CI tests before submission. \\
\addlinespace[2pt]
Data science / AI electives & Students compare model outputs across two prompting or modeling strategies, analyze errors, and document the limits of the model. \\
\addlinespace[2pt]
Cybersecurity & Students evaluate AI-generated code for vulnerabilities, misuse risks, and insecure assumptions. \\
\addlinespace[2pt]
Capstone & Teams build an AI-assisted workflow or agent system and submit an architecture diagram, test plan, governance report, and demo. \\
\bottomrule
\end{tabular}
\end{table}

These updates depend on faculty readiness, which is uneven: instructors report wide variation in confidence and policy around generative AI, and many lack a shared model of responsible student use \cite{zastudil2023,denny2024}. A low-cost remedy is one shared AI-use policy plus a recurring per-term norming session in which faculty co-grade a few anchor artifacts to calibrate rubrics---building the grading consistency that the workload-aware assessment below requires.

\section{Workload-aware assessment}\label{sec:assessment}
Assessment validity is now the core bottleneck: when students can generate plausible code and prose quickly, courses must measure judgment, verification, and transfer rather than first-draft production---or ask more complex, integrative questions to begin with. The useful patterns are familiar: commit-history checkpoints and test-driven milestones expose reasoning and debugging; AI-disclosure rubrics separate assistance from understanding; short oral defenses verify ownership; portfolios with reflection memos measure transfer; and role-bounded agent logs preserve accountability in team projects.

\paragraph{Match assurance effort to risk.} Each validity gain costs faculty effort, and naive redesign produces either oversimplified autograded tasks or unsustainable manual review. The workable equilibrium matches assurance to risk---automate broad coverage, sample expensive checks, and reserve deep verification for where it matters most (Table~\ref{tab:assessmentworkload}).

\begin{table}[ht]
\centering
\caption{A workload-aware assessment mix: match assurance effort to risk.}
\label{tab:assessmentworkload}
\small
\begin{tabular}{@{}p{0.30\linewidth}p{0.30\linewidth}p{0.30\linewidth}@{}}
\toprule
\textbf{Component} & \textbf{Coverage} & \textbf{Workload strategy} \\
\midrule
Milestones and tests & All students & Automated \\
\addlinespace[2pt]
AI disclosure & All students & Lightweight rubric \\
\addlinespace[2pt]
Oral defense & 20--40\% sampled & High-impact verification \\
\addlinespace[2pt]
Portfolio & 1--2 times per term & Batch grading \\
\addlinespace[2pt]
Logs and checkpoints & Team projects only & Structured summaries \\
\bottomrule
\end{tabular}
\end{table}

To make the section usable rather than conceptual, we give the actual mechanisms instructors can adopt directly.

\paragraph{AI-disclosure template.} A short, standard form students complete with each submission:
\begin{itemize}[noitemsep,topsep=2pt]
  \item Which AI tool did you use?
  \item What did you ask it to do?
  \item Which parts of the output did you accept, reject, or modify?
  \item How did you verify the result?
  \item What do you personally understand now that you did not understand before?
\end{itemize}

\paragraph{Oral-defense questions.} A two-to-three-minute sampled defense can use a small fixed bank:
\begin{itemize}[noitemsep,topsep=2pt]
  \item Explain one design decision you made.
  \item Show one bug or weakness you found in the AI output.
  \item Walk through one test case.
  \item What would fail if the input changed?
  \item Which part of the work should not be trusted without human review?
\end{itemize}

\paragraph{Commit history and checkpoints.} When reviewing version history or milestones, instructors look for evidence of incremental progress, student-written explanations, debugging attempts, tests added before final submission, and changes made after AI-generated drafts.

\paragraph{Portfolio assessment.} A portfolio submission should include the final artifact, a reflection memo, an AI-use disclosure, verification evidence, lessons learned, and a short statement of limitations and risks.

Crucially, these redesigns change the \emph{nature} of the work, not just its amount: a department invests once in reusable rubrics, disclosure templates, and autograding, then reuses them, shifting faculty effort from grading output volume toward evaluating thinking, process, and ownership. The tradeoff is worthwhile when it makes cheating less actionable and grading judgments more defensible.

\section{Workforce and stackable-credential pathway}\label{sec:workforce}
The same Capability Ladder competencies that structure a degree pathway also define a workforce-upskilling pathway. Sectoral, employer-driven training---aligning curricula to industry skill needs and connecting completers to jobs---produces some of the most credible earnings gains in the workforce literature, including from randomized evaluations \cite{katz2022}, and short, stackable credentials show positive labor-market returns \cite{soliz2023} and have been implemented within computing programs so that short units ladder into degree credit \cite{chakravorty2023}. Because the ladder is tool- and domain-agnostic and reachable with low/no-code platforms, the same units package as both degree courses and short, employer-validated credentials for incumbent and career-changing professionals in any AI-adopting industry (e.g., healthcare, finance, advanced manufacturing). A learner might complete L0--L2 units as a non-credit ``AI workflow'' micro-credential, then ladder the same competencies into degree credit later---lowering the barrier to entry while preserving a path to deeper study.

\section{An exploratory pilot course}\label{sec:pilot}
We developed an upper-division, cross-listed computing/business elective at a teaching-focused public university---with no programming prerequisite---and ran it over two semesters (Fall~2025 and Spring~2026). We report it as a preliminary, single-program experience, not a controlled study, and we draw design lessons rather than causal conclusions. The course ran as regular instruction; the informal follow-up below was instructor outreach, not research under an IRB protocol, so we present it as anecdote rather than human-subjects findings.

\paragraph{Design and methods.} The course was an \emph{elective} open to upper-division students, enrolling a mix of computing-major and business/non-CS students (about 40 per offering, roughly 80 in total). Rather than a content-delivery model, it taught students to use AI creatively to frame, test, and solve authentic problems. Students formed cross-functional teams---deliberately mixing computing and business backgrounds---and worked across the Capability Ladder on a commercial no-code agent-orchestration platform (named withheld for anonymity) to tackle real problems contributed by regional industry partners. Each student took a defined ``AI role'' (e.g., workflow architect, agent supervisor, verification/QA lead, or AI project manager), supported by a lean build methodology, entrepreneurship mentorship, and a demo day. Business-track students emphasized scoping, workflow design, customer discovery, and communication, while computer-science students emphasized technical integration, verification, and implementation; cross-functional learning came from coordinating across roles and defending joint decisions. Assessment emphasized role-specific evidence---students were evaluated on the capabilities relevant to their assigned role rather than on identical programming outputs---and produced employer-facing portfolios. Table~\ref{tab:pilot} summarizes the design.

\begin{table}[ht]
\centering
\caption{Pilot course at a glance.}
\label{tab:pilot}
\small
\begin{tabular}{@{}p{0.22\linewidth}p{0.40\linewidth}p{0.30\linewidth}@{}}
\toprule
\textbf{Course feature} & \textbf{Implementation} & \textbf{Evidence collected} \\
\midrule
Audience & Upper-division elective, no programming prerequisite; mixed CS and business students & Enrollment records \\
\addlinespace[2pt]
Structure & Cross-functional teams; defined AI roles; lean build + demo day & Team and role assignments \\
\addlinespace[2pt]
Platform & No-code agent-orchestration platform; work across the Capability Ladder & Student-built workflows/agents \\
\addlinespace[2pt]
Projects & Real problems from regional industry partners & Employer-facing portfolios \\
\addlinespace[2pt]
Assessment & Role-specific evidence; verification of AI output & Portfolios, demos, reflections \\
\bottomrule
\end{tabular}
\end{table}

\paragraph{Observations, stated cautiously.} The structure forced \emph{verification and supervision} of AI output on authentic tasks. Notably, business-track students reached the agent and agent-team levels without prior programming coursework---suggesting that the no-code ladder widens rather than restricts participation.
The follow-up contact group was a minority of the roughly 80 enrollees, self-selected, and not systematically tracked, so outcome reports from it are anecdotes, not rates. Within that group, approximately one-fifth later reported involvement in related AI projects, internships, or roles, and working across varied domains was associated with four follow-on projects. We do not claim the course caused these outcomes, only that they were reported after it.

\section{A staged adoption roadmap}\label{sec:roadmap}
The framework is not a single multi-year rollout; it supports three adoption speeds, each with concrete deliverables, so a department can start now while larger change matures.

\paragraph{Immediate (one semester).} In a single course, produce: one shared AI-use policy; one disclosure template; one assignment redesigned to allow AI but require verification; one sampled oral-defense protocol; and one simple rubric for verification. No new infrastructure is required.

\paragraph{Program (two to three semesters).} Across several courses, produce: a shared Capability-Ladder vocabulary; common rubric language; two or three courses with aligned AI-use expectations; sample student artifacts for calibration; and a faculty norming session each term.

\paragraph{Institutional and workforce (one to two years).} At program scale, produce: an employer-partner project bank; short credential modules; portfolio requirements; advisory-board review; and a placement or employer-feedback loop.

None of these moves is novel in isolation; many departments already use several of them \cite{shein2024,denny2024}. The roadmap's contribution is the sequencing: shared ladder language lets single-course changes compound into program-level alignment. Across all three, change scales with signal strength rather than headlines, governed by per-term trigger thresholds (oral-check anomaly rates, capability-attainment benchmarks, employer feedback) and kept low-overhead through shared templates and one standardized agent-use rubric.

\section{Limitations and risks}\label{sec:limits}
Several limits bound the claims above:
\begin{itemize}[noitemsep,topsep=2pt]
  \item This is a structured narrative review, not a systematic review or meta-analysis.
  \item Labor-market evidence is changing quickly and may be confounded by macroeconomic factors, pandemic hiring corrections, tax policy, offshoring, and remote work.
  \item Industry reports are used as directional evidence, not causal proof.
  \item The pilot course was exploratory, single-institution, and not controlled; its follow-up was informal, self-selected, and not conducted under an IRB protocol.
  \item The Capability Ladder still needs validation across more institutions, course types, and learner populations.
\end{itemize}

\paragraph{Risks of no-code agent platforms, and safeguards.} Low/no-code platforms widen participation, especially for business students and workforce learners, but they also create risks: students may not understand what the system is doing internally; platform-specific skills may not transfer; debugging may be harder; privacy and data governance can be unclear; students may over-trust agent outputs; and vendor lock-in may threaten sustainability. These risks can be reduced by pairing the platform with assurance activities---requiring workflow diagrams, test cases, verification checklists, human-approval points, and reflection memos---so that ease of building never substitutes for understanding and verification.

\section{Conclusion}\label{sec:conclusion}
AI is reallocating computing tasks toward verification, orchestration, supervision, security, and systems thinking faster than many curricula have changed. The Capability Ladder gives departments a way to modernize without chasing every new tool. It helps faculty decide what students may automate, what they must still understand, how they should verify AI output, and what evidence demonstrates responsible supervision. That makes it useful both for degree programs and for short workforce credentials.

That value should outlast the moment: as AI roles and tools normalize over the next three-to-five years, the durable capabilities the ladder targets---building, verifying, and supervising automation---are what remain once specific tools are commoditized. The same model scales into a degree pathway and compresses into short, stackable, employer-validated credentials \cite{katz2022,soliz2023}, adoptable in small, workload-aware steps while preserving durable computing foundations.

\bibliographystyle{unsrtnat}
\bibliography{references}

\end{document}